\documentclass[letterpaper, 10 pt, conference]{ieeeconf}  % Comment this line out
\IEEEoverridecommandlockouts                              % This command is only
\usepackage{graphics} % for pdf, bitmapped graphics files
\usepackage{epsfig} % for postscript graphics files
\usepackage{amsmath} % assumes amsmath package installed
\usepackage{amssymb}  % assumes amsmath package installed
\usepackage{array}

\usepackage{cite}

\newcommand{\Rmnum}[1]{\uppercase\expandafter{\romannumeral#1}}

\usepackage{graphicx}
\usepackage{subfig}
\usepackage{float}
\usepackage{algorithm}
\usepackage{algorithmic} % error

\floatname{algorithm}{Algorithm}

\title{\LARGE \bf
A Robotic Auto-Focus System based on Deep Reinforcement Learning*
}

\author{Xiaofan Yu, Runze Yu, Jingsong Yang, Xiaohui Duan.% <-this % stops a space
\thanks{*This work was supported by PKU-AMSKY joint innovation lab of intelligent equipments.}% <-this % stops a space
\thanks{The authors are with the Center of Wireless Communication and Signal Processing, Peking University, Beijing, 100871, China.
        {\tt\small Email: duan@pku.edu.cn}}%
}

\begin{document}

\maketitle
\thispagestyle{empty}
\pagestyle{empty}

%%%%%%%%%%%%%%%%%%%%%%%%%%%%%%%%%%%%%%%%%%%%%%%%%%%%%%%%%%%%%%%%%%%%%%%%%%%%%%%%
\begin{abstract}

Considering its advantages in dealing with high-dimensional visual input and learning control policies in discrete domain, Deep Q Network (DQN) could be an alternative method of traditional auto-focus means in the future.
In this paper, based on Deep Reinforcement Learning, we propose an end-to-end approach that can learn auto-focus policies from visual input and finish at a clear spot automatically. We demonstrate that our method - discretizing the action space with coarse to fine steps and applying DQN is not only a solution to auto-focus but also a general approach towards vision-based control problems. Separate phases of training in virtual and real environments are applied to obtain an effective model. Virtual experiments, which are carried out after the virtual training phase, indicates that our method could achieve 100\% accuracy on a certain view with different focus range. Further training on real robots could eliminate the deviation between the simulator and real scenario, leading to reliable performances in real applications.

\end{abstract}

%%%%%%%%%%%%%%%%%%%%%%%%%%%%%%%%%%%%%%%%%%%%%%%%%%%%%%%%%%%%%%%%%%%%%%%%%%%%%%%%
\section{INTRODUCTION}
% using vision input
Automatic focus is the first and foremost step in cell detection using microscopic images, which has wide applications in diagnoises of diseases like tuberculosis \cite{saini2016comparative,mateos2012comparative}. Auto-focus mechanisms are divided into two categories: active auto-focus and passive auto-focus \cite{chen2010passive}. In this paper, we concentrate on passive auto-focus problems that aim at moving the lens to a clear enough view within acceptable time by analyzing captured images. Solutions to passive auto-focus problems have two phases: the focus measure functions that map an image to a value for representing the degree of focus of the image, and the search algorithms that iteratively move the lens to find the highest or nearest peak of focus measure curves \cite{mir2015autofocus,li2005autofocus}. In order to improve traditional passive auto-focus techniques, both of the phases need to be considered. In this paper, we present an end-to-end learning approach by combining those two phases into one. \par

Aiming at giving agents the ability to learn directly from raw pixels, we need methods which could learn to make decisions on a model-free problem with high-dimensional input. Deep Reinforcement Learning (DRL), which combines Deep Learning and Reinforcement Learning (RL) together, is promising for complex decision-making tasks \cite{arulkumaran2017brief} and thus becomes a potential solution. One typical example of DRL is the Deep Q Network (DQN), which has achieved performances that are compatible or even superior than human players in Atari games \cite{mnih2013playing,mnih2015human}. The basic idea of our system comes from formulating the auto-focus problem into a model-free decision-making task analogous to Atari games: we formulate the state representation with successive images and limit the optional actions to five, including coarse and fine step in both directions as well as termination. \par

%By doing this, we intend to utilize DQN's ability of handling high-dimensional data and investigate the possibility of learning focus policies through interactions. 
The contribution of this paper can be summarized as follows:
\begin{enumerate}
  \item We address the problem of auto-focus by applying vision-based DRL, which does not utilize any human knowledge and therefore is superior to supervised learning in achieving superhuman performances \cite{silver2017mastering}. To the best of our knowledge, this is the first demonstration of microscopic focusing that uses vision input and learns through trial and error. 
  \item Our paper demonstrates a general approach to vision-based control problems by discretizing the action space to apply coarse-to-fine strategies. Our model is trained in virtual environment first and real world after that. \par
\end{enumerate}

% As first steps, we assess the feasibility of our model on a self-built general platform, as shown in Fig. 1. Our system consists of an optical table, a XSZ-4GA optical microscope and a UR3 robotic arm with its gripper. The position of the microscope and the initial position of the robotic arm is mannually configured in advance. We use the robotic arm to screw the focusing knob as an alternative to the motors in specialized machines, which brings the advantage in adaptability and easy collabrations with human. We evaluate our trained model in both the virtual and real scenarios, from the aspect of accuracy and speed, which are the most crucial part to user satisfaction \cite{mir2015autofocus}.\par 

% paper structure 
% 3: formluate the robotic reinforcement learning problem, introduce essential notation and describe the exisiting algoritmatic foundations on which we build the methods for this work. @ Deep reinforcement learning for robotic manipulation
The remainder of this paper is organized as follows: Section \ref{sec:related} introduces related works. Section \ref{sec:method} formulates the auto-focus RL problem and presents an overview of our method. Section \ref{sec:experiments} describes the experimental setup and the experiment results in both virtual and real scenarios. \par

%%%%%%%%%%%%%%%%%%%%%%%%%%%%%%%%%%%%%%%%%%%%%%%%%%%%%%%%%%%%%%%%%%%%%%%%%%%%%%%%
\section{RELATED WORK}
\label{sec:related}

In this section, we describe related works in passive auto-focus and vision-based DRL for control. \par

\subsection{Passive Auto-Focus}

As described above, existed passive auto-focus techniques contain two separate sections: a focus measure function and a search algorithm. Previous works contributed to these sections separately. Comparative works were done in evaluating different focus measure functions, e.g. Gaussian Based Methods, Laplacian Based Methods, etc., in execution time, correctness and other critical criterias \cite{saini2016comparative,mateos2012comparative}. In this paper, we choose Tenengrad method (one type of Gaussian Based Methods) as a reference for reward due to its balanced performace on execution time and correctness. As for search algorithms, traditional methods focused on finding the nearest peak \cite{he2003modified} or the highest peak and all peaks \cite{kehtarnavaz2003development} based on intuitive ideas. Recent research utilized neural networks to search and achieved less computational complexity \cite{chen2010passive} as well as faster speed \cite{mir2015autofocus}, but they still applied focus measure values as inputs to feed the neural network. \par

% nearest peak
% Previous passive auto-focus approaches did successfully focus in many cases. However, they rely heavily on the focus measure functions, which have the drawback of focusing on inconsequential objects and thus generating local maximas \cite{mir2015autofocus}. Too many local peaks could lead the focus process to failure. In this paper, we take first steps towards using raw pixels instead of focus measure functions in the forward calculation to avoid this problem. \par

Contrary to previous techniques, our method is an end-to-end approach which directly processes vision input and outputs an action decision. By combining the two distinct phases into one, our approach enables the agent to complete ``perception'' and ``control'' as a whole.

\subsection{Vision-Based DRL for control}

In vision-based DRL, approaches towards problems with discrete and continuous action spaces are largely different \cite{metz2017discrete}. As a classical function approximation method, DQN is widely applied in discrete domain such as games with limited actions \cite{mnih2013playing,mnih2015human,silver2017mastering}. Policy search methods, on the other hand, provide solutions for robotic control problems which require continuous instructions to set motor torques \cite{levine2016end,schulman2015trust}. However, direct policy search methods have inherent difficuties in maximization and backup calculations for its non-negligible complexity \cite{metz2017discrete}, which results in recent works on utilizing value-function-based update in continuous action problems. Normalized Advantage Functions (NAF) made successful attempts in extending the Q-learning variant to continuous domain\cite{gu2016continuous}. Another approach, called Deep Deterministic Policy Search (DDPG), circumvented the problem by adopting actor-critic methods \cite{lillicrap2015continuous}. These means learned promising policies in continuously-controlled physical tasks, but the dimension of input data is strongly limited \cite{gu2017deep}. To summarize, DQN is still the first choice for discrete action problems considering its advantage in processing high-dimensional input. \par

% RL stands for a promising solution to complex decision-making tasks such as playing games \cite{silver2017mastering,kinoutnik2014evolving}, autonomous vehicles control \cite{zhang2016learning} and robotic {}manipulations \cite{levine2016learning} by simply providing reward information. When combining RL and DL by applying Convolutional Neural Network (CNN) to the agent, DRL is able to process high-dimensional vision input and thus open up the possibilities of automating more complex physical tasks without prior knowledge \cite{zhang2015towards,lampe2013acquiring}. \par

In our paper, aiming at directly processing vision input as well as successfully focusing with limited complexity, we discretize the action space and apply DQN on auto-focus, which is detailedly described in Section \ref{sec:method}. Our method is unique in two aspects: first, we formulate the auto-focus problem to a discrete decision-making task with specially designed state representation and action space. Second, a reward function including an active termination mechanism is presented based on focus measure functions. The key ideas of our approach could be applied to other visiual feedback control applications.

%%%%%%%%%%%%%%%%%%%%%%%%%%%%%%%%%%%%%%%%%%%%%%%%%%%%%%%%%%%%%%%%%%%%%%%%%%%%%%%%
\section{METHOD}
\label{sec:method}

In this section, we define the auto-focus RL problem and present an overview of our approach. We will explain our auto-focus system based on RL from system model, reward function design and DQN design.

\subsection{System Model}

% state, action, reward(section 3.2), formulars(Loss function, update functionz) 
The basic principle of RL can be demonstrated as a loop: the agent selects an action ${a_t}$ from a designed legal action set ${\rm A}$ and passes it to the environment. The environment triggers a state transition from ${s_t}$ to ${s_{t+1}}$, returning with a reward ${r_t}$ and a new observation ${x_t}$. The goal of the agent is to learn a policy that can maximize the future rewards ${R_t} = \sum\limits_{t' = t}^T {{\gamma ^{t'-t}}{r_{t'}}}$, where $\gamma$ is a discount factor and T is the terminal timestep. During the learning process, the Q function $Q(s,a)$ which stands for an evaluation of the current policy, is updated by the \emph{Bellman Equation} at each iteration ${i}$:
\begin{equation}
{Q_{i + 1}}(s,a) = {\mathbb{E}_{s'}}\left[ {r + \gamma {{\max }_{a'}}{Q_i}(s',a')|s,a} \right] \label{func:bellman}
\end{equation}
where $s'$ and $a'$ are the next state and next action based on $s, a$. Obtaining an optimal Q function ${Q^*}(s,a)$ that describes the best policy is our goal. \par

\begin{figure}[t]
  \centering
  \includegraphics[width=0.35\textwidth]{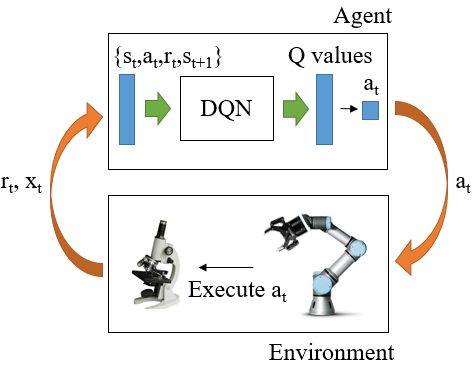}
  \caption{System model of auto-focus based on RL in real environment.}
  \label{img:systemmodel}
\end{figure}

\begin{figure*}[t]
  \centering
  \includegraphics[width=0.9\textwidth]{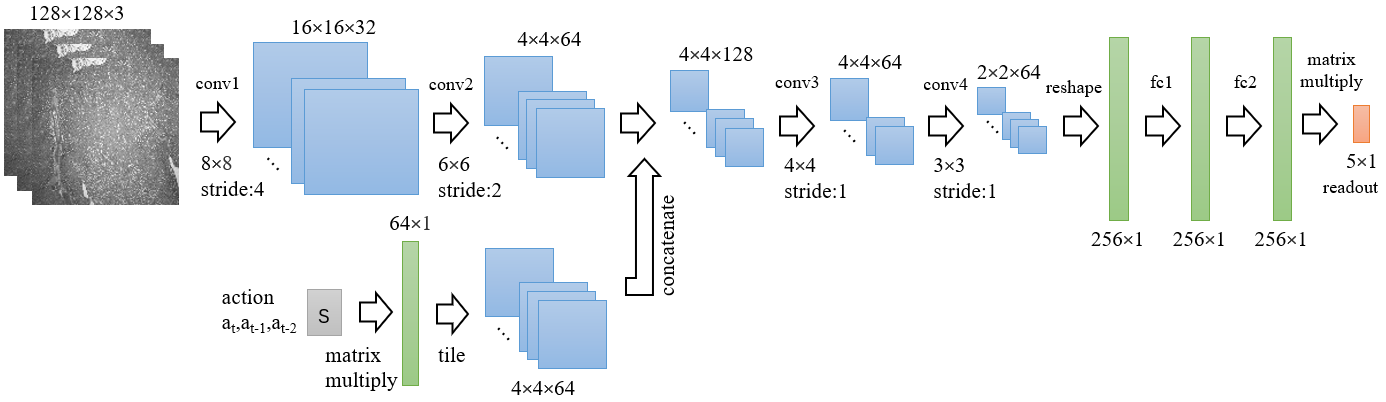}
  \caption{The architecture of our DQN.}
  \label{img:DQNstructure}
\end{figure*}

% unique property of the problem itself
% The problem of auto-focus is unique in the requirement of active termination and a trade-off between speed and accuracy.  Besides, taking human visual system limitations into account \cite{li2005autofocus}, it is unnecessary to find the clearest position and thus discretization upon lens' positions is reasonable. The step of discretization should be determined carefully: too large steps may preclude an acceptable position from the state space while too small steps is unnecessary and leads to time-consuming searches. The goal of our approach is to find a clear enough spot for human vision within acceptable time.

Important factors in DRL include state space, action space, reward function and Q function. In our approach, three successive images and their corresponding actions are packed together to a sequence as state representation: ${s_t} = \left\{ {{x_t},{a_t},{x_{t-1}},{a_{t-1}},{x_{t-2}},{a_{t-2}}} \right\}$.
The action space ${\rm A}$ is set with 5 discrete options: coarse positive step, fine positive step, terminal step, fine negative step and coarse negative step. Our design of the discrete action space reveals the idea of coarse-to-fine strategy, which is an old but very common thought in auto-focus and control problems \cite{ren2008vision,ralis2000micropositioning} by converging fast as well as accurately to the best solution. The reward function and Q function (DQN) are the central part of our method, thus are demonstrated separately in Section \ref{sec:reward} and \ref{sec:DQN}. \par 

The system model of our auto-focus platform is shown in Fig. \ref{img:systemmodel}. In real environment, our system learns through interactions: a chosen step is performed to the environment, after which the environment returns with a reward $r_t$ and a new observation $x_t$ - the real-time image obtained from the microscope. Updated with ${x_t}$ and ${a_t}$, the state sequences are fed to DQN, which calculates Q functions and outputs the Q values regarding each action. The action with the maximum Q value is selected as the next motion, which will open up the subsequent iteration. Such a model works on any vision-based problem.
% discrete!  Our method reduces the complexity of the problem, increases the processing speed as well as limits the loss in accuracy to human tolerance.

% The kernel flow of the reward function algorithm is shown in Algorithm 1.

%\begin{algorithm}[h]
%	\caption{Reward Function}
%		\begin{algorithmic}[1]
%		\REQUIRE ~~\\
%				 $s_t$: current state \\ % a sequnce of previous images and actions 
%				 $a_t$: input action \\
%				 $step$: the number of current step \\
%				 $cur\_angle$: current angle of the focusing knob 
%	    \ENSURE  ~~\\
%	    		 $s_{t+1}$: next state \\
%	    		 $r_t$: reward \\
%	    		 $t$: whether the auto-focus is terminal
%	    \STATE ${s_{t + 1}} = StateTransfer({s_t},{a_t})$
%	    \STATE $CurFocus = ComputeFocus({s_{t + 1}})$
%	    \STATE $Dis = CurFocus - MaxFocus$
%	    \IF {$cur\_angle < MIN\_ANGLE$ \OR $cur\_angle < MAX\_ANGLE$}
%	    \STATE $r_t = FAILURE\_REWARD; t = TRUE$
%	    \ELSIF {$step > MAX\_STEP$}
%	    \STATE $r_t = FAILURE\_REWARD; t =$ \TRUE
%	    \ELSIF {$a_t == TERMINAL$}
%	    	\IF {$CurFocus > 0.9 \times MaxFocus$}
%	    	\STATE $r_t = SUCCESS\_REWARD; t =$ \TRUE
%	    	\ELSE
%	    	\STATE $r_t = FAILURE\_REWARD; t =$ \TRUE
%	    	\ENDIF
%	    \ELSE
%	    \STATE $r_t = Dis$
%	    \STATE $t =$ \FALSE
%	   	\ENDIF
%    \end{algorithmic}
%\end{algorithm}
\subsection{Reward Function Design}
\label{sec:reward}

The most important part of our system is the reward function for the following reasons. First and foremost, it determines the reward ${r_t}$ and guides the learning process: when returned with a high reward, the agent tends to perceive the previous action as a good policy. An inadequate reward may result in a failed training or largely impede the training speed. Second, the reward function is also crucial in providing terminating conditions. In this paper, we use the absolute difference between current and maximum focus measure values to determine rewards, which is simple and intuitive but may cause chaos when the focus measure curve displays too many local peaks. To address this problem, we mannually control the region-of-interest to a rich-content area. As for terminating conditions, unlike any of the control problems defined in \cite{riedmiller2005neural}, auto-focus does not have an objective judgement that is offered by the environment. Instead, the agent is desireable to stop actively once satisfies the conditions. In our approach, the following situation is considered as a success: actively choosing to terminate at a position where the focus measure value is greater than 0.9 times of the maximum value in this field. Other cases such as moving beyond the legal range, excessing maximum steps or stopping at a blur vision will all be regarded as failure and given a large negative reward. Our reward function could be described in (\ref{fun:reward}), where ${t}$ is ${100}$ or ${-100}$ depending on success or not and ${c}$ is simply a coefficient.
\begin{equation}
reward=c \cdot (cur\_focus-max\_focus)+t \label{fun:reward}
\end{equation}

\subsection{DQN Design}
\label{sec:DQN}

The Q function is defined by DQN, which parameterizes the optimal Q function by Convolutional Neural Network (${Q^*}(s,a) \approx Q(s,a;\theta )$) and updates these weights by minimizing the loss function ${L_i}({\theta _i})$:
\begin{equation}
{L_i}({\theta _i}) = \mathbb{E}\left[ {{{({y_i} - Q(s,a;{\theta _i}))}^2}} \right] \label{func:loss}
\end{equation}
where ${y_i} = {\mathbb{E}_{s'}}\left[ {{r_t} + \gamma {{\max }_{a'}}Q(s',a';{\theta ^{i - 1}})|s,a} \right]$ is the target for iteration ${i}$. \par

The structure of our DQN is shown in Fig. \ref{img:DQNstructure}. The input images, after resizing and graying, are fed into 4 convolutional layers (conv). Each layer contains a convolution, a batch normalization, a rectified linear unit, followed by a max-pooling. With reference to the network structure in \cite{levine2016learning}, we transform the previous action sequences to a vector that has the same size as the vector originated from the input images, after which a concatenation of these two vectors is performed right in the middle of our network. The result of convolutional layers is then processed by two fully connected layers (fc) with 256 units. Finally the network outputs the Q value of each action through a matrix multiplication. In all, our DQN directly processes high-dimensional state sequences and actions while outputs Q values of each valid action. \par

The whole network contains 381K parameters and requires 13.8M multiply-accumulate operations in each update.
% The action sequences are added into the network after a matrix multiplication and a tiling.
%%%%%%%%%%%%%%%%%%%%%%%%%%%%%%%%%%%%%%%%%%%%%%%%%%%%%%%%%%%%%%%%%%%%%%%%%%%%%%%%
\section{EXPERIMENTS}
\label{sec:experiments}

In this section, we present our auto-focus experiments from the following aspects: hardware setup, training in virtual environment and training in real environment.

\begin{figure}[!b]
  \centering
  \includegraphics[width=.45\textwidth]{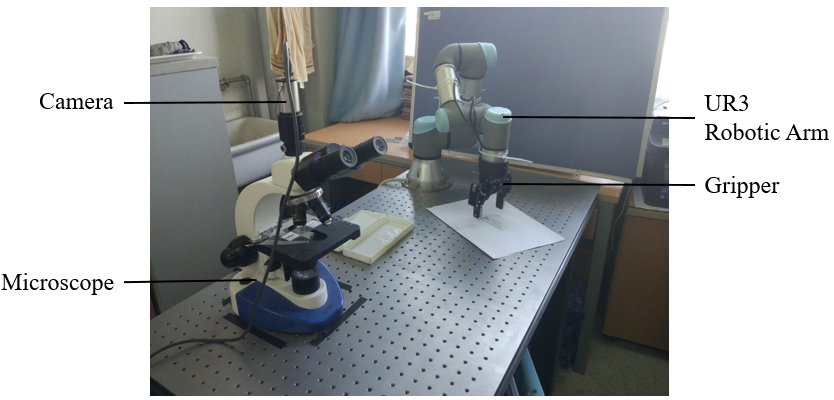}
  \caption{Auto-focus system implementation of the optical table, the microscope and the robotic arm. Another part of our system is the PC, which locates out of the right boundary of this figure.}
  \label{img:realsystem}
\end{figure}

\subsection{Hardware Setup} % hardware and test index

We build a general system with a robotic arm to evaluate our algorithm. Our platform has two distinct sections: training in virtual and in real environment. Before training in simulation, we collect a series of images at equal-spaced positions and generate an auto-focus simulator, where the agent learns through interacting with it. For training on real robot, our system can be categorized into three parts: a PC where settled the DQN, a robotic arm with its gripper and a microscope with a camera on its eyepiece. Sending instructions from the PC, we use the robotic arm to execute the action $a_t$ by screwing the focusing knob as an alternative to the motors in specialized machines. This setting brings the advantage in adaptability and easy collabrations with human. Controlled by the computer, the microscopic camera is responsible for sending observation images $x_t$ to the PC. \par

As for hardware setup, our system consists of a personal computer, an optical table, a XSZ-4GA optical microscope and a UR3 robotic arm with its gripper. Part of the implementation is shown in Fig. \ref{img:realsystem}. Important parameters of our hardware is listed as follows: the personal computer, with 8 Intel(R) Xeon(R) E5-1620 @3.50GHz CPUs, a 64GB RAM and 2 NVIDIA Quadro M4000 8GB GPUs, is the host of the whole system including DQN and interface modules. The optical table serves as a stable foundation of the microscope and the robots. UR3 screws the focusing knob with a minimum step of 0.001 rad. The QHY microscopic camera works on a $1944 \times 2592$ resolving power, taking single frame each iteration. \par

%Currently, we connect those components with each other through wires. % the PC sends intructions to and requires state information from the robotic arm through an Ethernet cable while it connects to the gripper through a USB-RS485 converter. The microscopic camera is joined with the PC through a USB cable. \par

\subsection{Training in Virtual Environment}

Considering the huge time-consumption of operating a real robot, we add a virtual training step in a simulator before applying our DQN to real scenario. To construct the simulator, a collection of the eyepiece's views with equal spacing across the full ``focus range'' is carried out. The focus range is defined by absolute angles of the focusing knob, using the unit of radian. With the simulator, the focusing process can be simulated by performing discrete actions upon an observation and tranfering between stored images. One thing to note is that, it is the discrete action design that enables the easy implementation of training in virtual environment. \par

By applying DQN to the simulator, we evaluate the feasibility of our algorithm and its adaptability to broader focus ranges as well as different views in 3 sets of experiments. These experiments, with their setups shown in Table \ref{table:experimentsetup}, will be discussed individually in the following sections. In virtual tests, we evaluate training models from cost and accuracy, which refer to the cost in (\ref{func:loss}) and the success rate on test set respectively. The cost are recorded as the training processes, while the accuracy comes from a simutaneous evaluation on the same data set every 1K timesteps. An effective learning approach should end up with a convergent cost and a 100\% accuracy. Or in other words, the smaller cost and the larger accuracy, the better. An extra virtual test is performed after training to evaluate the accuracy and speed of that trained model. \par

Some static settings of the experiments are summarized as follows: in our approach, every episode starts with a random position among the discretely-collected ones while the position of the microscope and the initial posture of UR3 are mannually configured in advance. The absolute angular change of a coarse step and a fine step are 2.7 rad and 0.3 rad respectively while the permitted maximum step is 20. \par

\subsubsection{Basic experiment} 

\begin{figure}[t]
  \centering
  \subfloat[Result of experiment 1: training on 30.0-69.0 rad with the same view as train and test data set. Cost is displayed leftwards while accuracy is on the right.]{\includegraphics[width=0.45\textwidth]{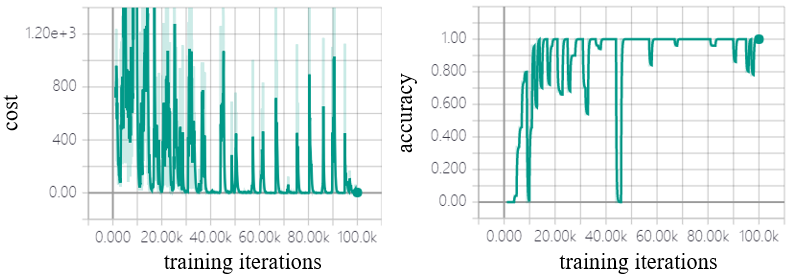} \label{img:vexp1}} \\
  \subfloat[Result of experiment 2: training on 10.2-89.7 rad with the same view as train and test data set. Cost is displayed leftwards while accuracy is on the right.]{\includegraphics[width=0.45\textwidth]{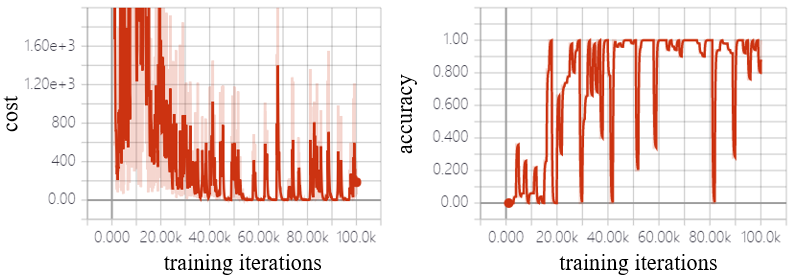} \label{img:vexp2}} \\
  \subfloat[Result of experiment 3: training on 30.0-69.0 rad with different views as train and test data set. Test accuracy on the similar field is displayed leftwards while accuracy on the different slice is on the right.]{\includegraphics[width=0.45\textwidth]{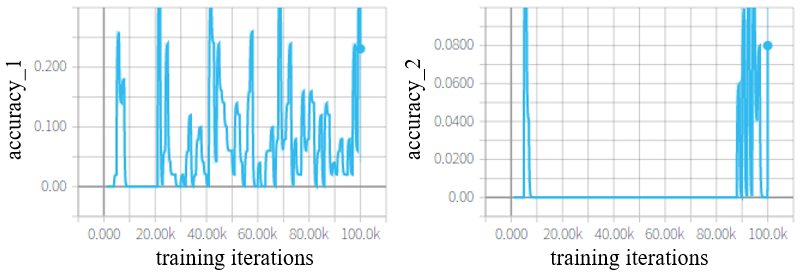} \label{img:vexp3}} \\
  \caption{Results of virtual experiments, with x-axis representing training iterations while y-axis describing the value of cost or accuracy.}
\end{figure}

The first experiment is for assessing the feasibility of our method, so we use the same view as training and testing set with a middle-sized focus range: not too broad but very enough to generate a hill-like focus measure curve. The training process with 100K timesteps lasts 5hrs and 14mins, resulting in a learning curves shown in Fig. \ref{img:vexp1}. \par

Due to random actions in $\varepsilon {\rm{ - }}greedy$ policy, the learning cost in Fig. \ref{img:vexp1} displays a trend of gradually decreasing but with many fluctuations. While jumping to 100\% after approximate 15K iterations of training, the accuracy converges slowly to 100\% after that. The extra test shows the model we learned after 100K times training can achieve 100\% accuracy with 5.611 steps in average, proving the feasibility of our method.

% make a table
\begin{table}[t]
  \caption{Experimental setups}
  \label{table:experimentsetup}
  \begin{center}
    \begin{tabular}{|l||p{2.5cm}<{\centering}||p{1.5cm}<{\centering}||p{2.25cm}<{\centering}|}
    \hline
    No. & Goal & Focus Range (rad) & Train \& Test Data Set \\
    \hline
    1 & Basic experiment to assess the feasibility & 30.0-69.0 & Same view \\
    \hline
    2 & Comparison experiment to assess the adaptablity to broader focus range & 10.2-89.7 & Same view \\
    \hline
    3 & Comparision experiment to assess the adaptablity to different views & 30.0-69.0 & Three different views, one for training and the rest two for testing \\
    \hline
    \end{tabular}
  \end{center}
\end{table}

\subsubsection{With broader focus range}

This experiment aims at testing the adaptability of our method when extending the legal focus range. To make a comparison, we use the same view as the one in previous experiment, only leaving the permitted focus range in difference. A training with 100K timesteps takes 6hrs and 32mins with the learning curves shown in Fig. \ref{img:vexp2}. Compared with learning curves in Fig. \ref{img:vexp1}, the accuracy curve here reaches 100\% slower and displays more dives, which is reasonable as a wider scope brings more difficulties in focusing. The addtional virtual test returns with 100\% accuracy and 8.578 steps averagely, indicating our method's outstanding adaptability to a broader focus range.

\subsubsection{Upon different views}

For auto-focus, the generalization capacity of an algorithm, or in other words, the performance when facing a brand new view that is different from the training set, is very critical for real-world applications. A strong generalization capacity means that to achieve the same accuracy and average steps, the amount of data needed to feed DQN is limited, thus the whole training process will be time-saving. \par

In this test, we apply three distinct fields as training and testing sets with other settings identical to experiment 1. One set is used in training and the rest two sets are testing sets that have varying degrees of similarities with the training set. The resulted training and testing accuracy curves are displayed in Fig. \ref{img:vexp3}, with the left graph representing using a similar field on one slice to test while the right graph standing for using two totally different fields from separate slices to train and evaluate. Although the training set achieves a high accuracy as usual, testing accuracy on both testing sets are poor, with an obvious better performance in the left figure. The test result shows that the generalization capacity of our method, which still needs to be improved, is strongly influenced by the similarities between training and testing microscopic views. \par

To summarize, our approach is able to successfully focus on certain views with fine adaptability to larger focus range and improvable generalization capacity in virtual training phase. Future works could be using various views to train in order to improve the generalization capacity.

\subsection{Training in Real Environment}

Based on the model learned in virtual experiment 1, we apply a similar training process on real robots and obtain a new model. We perform real tests on these two models separately to assess the effects of different training phases, which are described in the following sections. \par

In real tests, models are evaluated from the distribution of ``focused positions'', which is defined by the focus measure values of terminal positions. A good model has a larger possibility to end up at a clear view; that is to say, the better the model, the more the focused positions that locate at large values. By observing the distribution of focused positions, we can intuitively judge the performance of a certain model.

\subsubsection{Evaluate the virtually-trained model}

\begin{figure}[t]
	\centering
	\includegraphics[width=0.3\textwidth]{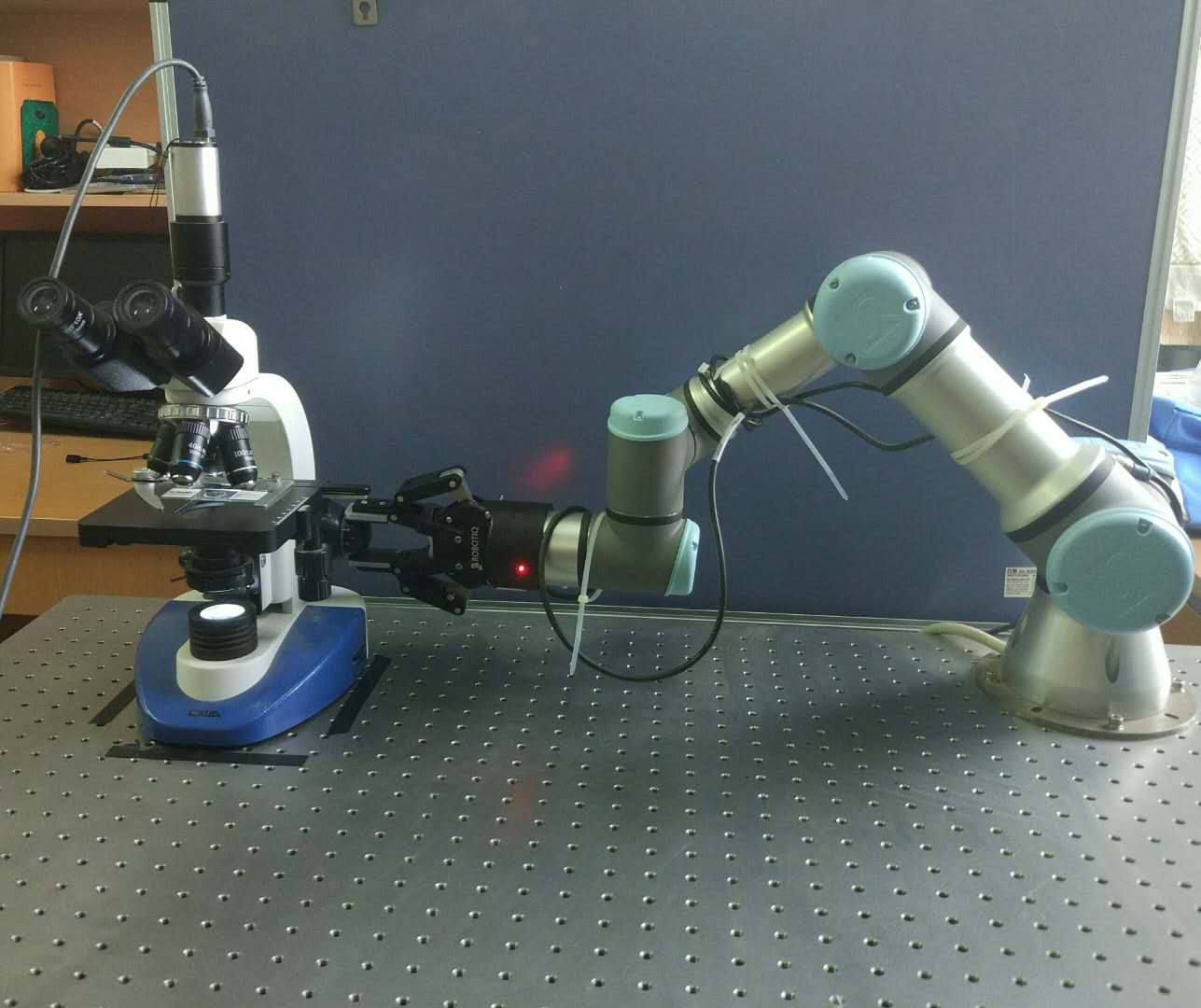}
	\caption{Real world testing scene.}
	\label{img:realscene}
\end{figure}

To evaluate the practical performance of the virtually-trained model, we simply load in that model, start a 10-times test on real robots and record the focused positions. Unlike the situations in virtual environment, every episode here starts with a random position in the continuous focus range rather than a pool of discrete positions. The rest settings including the microscopic view are identical to the conditions in virtual experiment 1. A scene of real world testings is shown in Fig. \ref{img:realscene}.

\begin{figure}[t]
	\centering
	\includegraphics[width=0.4\textwidth]{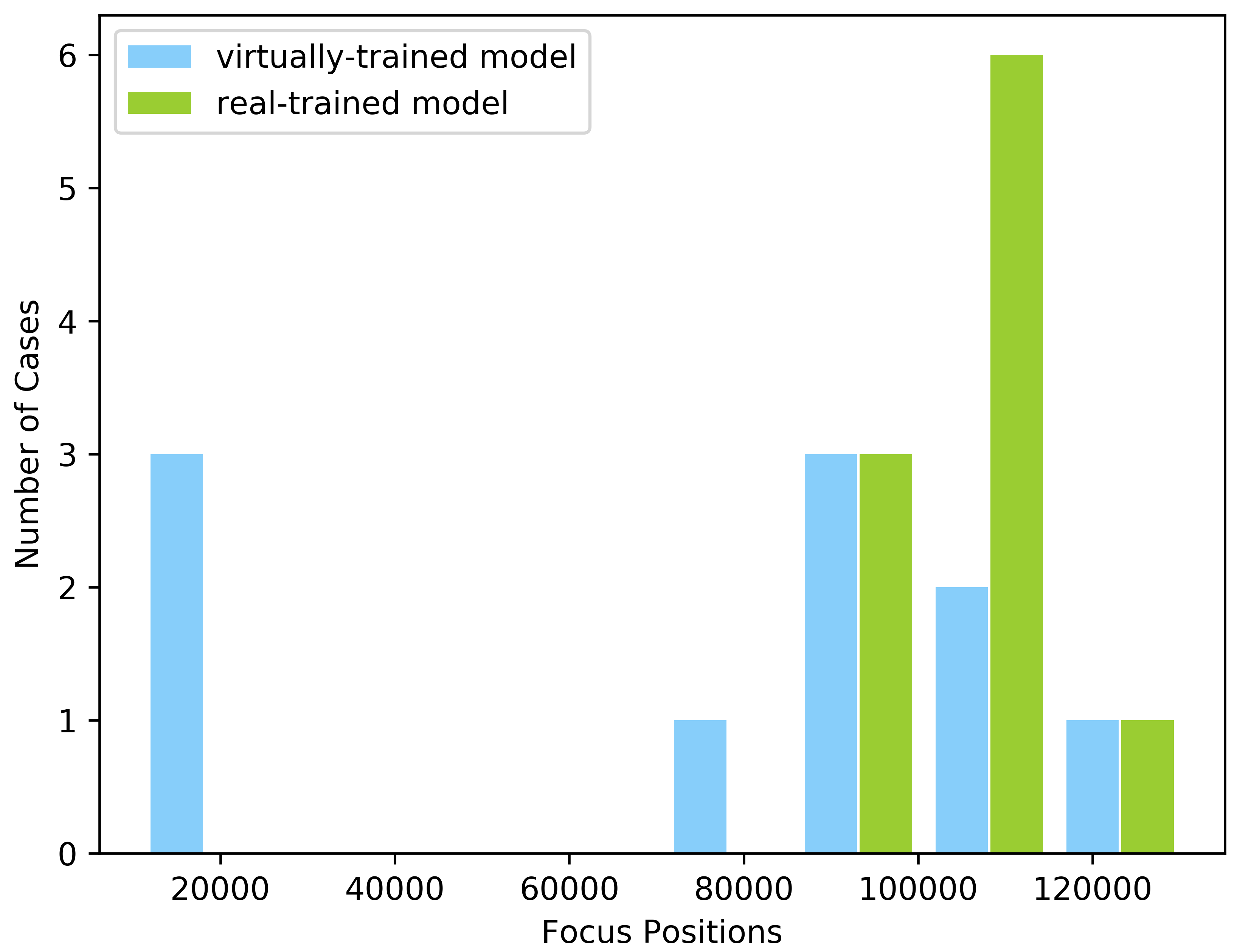}
	\caption{The histograms of focused positions, with x-axis representing focus positions while y-axis describing the number of cases that locate in corresponding bins.}
	\label{img:endff}
\end{figure}

The blue bars in Fig. \ref{img:endff} shows the test result of the virtually-trained model, from which we can conclude that our agent learns basic policy in virtual training but the policy is not accurate enough as 30\% of the cases result in serious failures. These failures may be attributed to the deviation between the simulator and real robots, e.g. knob slippage during screwing or a small shift of the microscopic view, which will be addressed by further training in real scenario.

\begin{figure}[t]
	\centering
	\includegraphics[width=0.47\textwidth]{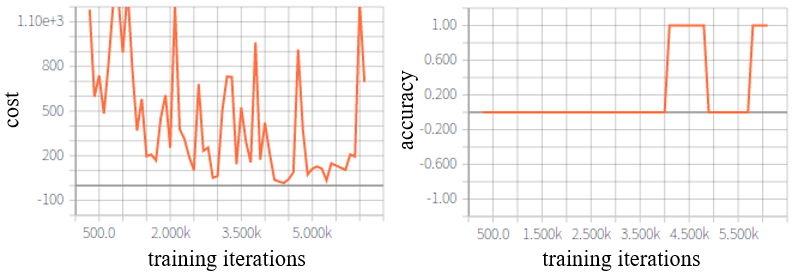}
	\caption{Results of training on real robots, with x-axis as training iterations and y-axis as cost or accuracy.}
	\label{img:realresult}
\end{figure}

\subsubsection{Evaluate the practically-trained model}

The real training phase runs for 6K timesteps within 6hrs and 22mins. As we expected, training in real scenarios is much slower than virtual phases due to the time-consuming robotic executions. Shown in Fig. \ref{img:realresult}, the testing accuracy jumps to 100\% after approximate 4K times training, indicating an adaption of environmental deviations. Such a small number in training episodes suggests the contribution of the virtual phase - saving much time in the real training stage. As for evaluation, a similar 10-times auto-focus test is performed on the platform with results shown as green bars in Fig. \ref{img:endff}. Compared to the distribution of the virtually-trained model, the focused positions of the practically-trained model move rightwards and all locate in a clear enough area for human vision, indicating a 100\% accuracy in real auto-focus. \par

Based on the test results above, it is reasonable to conclude that our method is feasible to learn accurate policies which can achieve 100\% accuracy on real equipments. However, the performance of our method is influenced by some environmental factors such as the shifting of field and thus addtional training phase is required to get adapt to these changes. The generalization capacity of our DRL auto-focus method and its robustness to environmental shiftings will be the key part in future improvements. 

%%%%%%%%%%%%%%%%%%%%%%%%%%%%%%%%%%%%%%%%%%%%%%%%%%%%%%%%%%%%%%%%%%%%%%%%%%%%%%%%
\section{CONCLUSIONS AND FUTURE WORKS}
\label{sec:conclusion}

In this paper, we present an end-to-end learning approach to auto-focus, which has important applications in diagnoises. We use DQN to process high-dimensional vision input and learn through interactions with the environment, also taking first steps in reward design by refering to focus measure values and adding an active terminal action. Besides, we demonstrate that discretization in the action space could be a general solution to all vision-based control problems. In the virtual experiments, our method shows great feasibility and adaptability to extended range but is relative weak when facing different views. Further training on real robot could eliminate the deviation between the simulator and real scenario, achieving reliable performances in real applications. \par

However, our method still has a long way to go before practical applications. Future works on reducing training time and improving performance, especially generalization capacity and robustness, are necessary. As is revealed in virtual experiment 3, the performance of our algorithm is much better when applied on analogous fields, which brings the possibility of improving the generalization capacity by training on larger datasets, e.g. collecting more fields on one slice. This idea is very similar to data augmentation. As for the robustness to environmental changes, addtional sensors could be a candidate by accurately tracking these variations. \par

It is also important to carry out comparison experiments between DRL algorithms with discrete and continuous action space. Continuous methods are more complex thus slower in speed but have advantages in accuracy. Auto-focus problem itself requires a trade-off between speed and accuracy, hence further tests of different approaches are desired to find the optimal one. Moreover, comparisons with traditional passive focus techniques are also necessary, from the aspect of correctness, speed, robustness to noise, etc.

% \addtolength{\textheight}{-3cm}   % This command serves to balance the column lengths
                                  % on the last page of the document manually. It shortens
                                  % the textheight of the last page by a suitable amount.
                                  % This command does not take effect until the next page
                                  % so it should come on the page before the last. Make
                                  % sure that you do not shorten the textheight too much.

%%%%%%%%%%%%%%%%%%%%%%%%%%%%%%%%%%%%%%%%%%%%%%%%%%%%%%%%%%%%%%%%%%%%%%%%%%%%%%%%
\section{ACKNOWLEDGMENTS}

The authors gratefully acknowledge the assistance on robot setup and advice on paper from Wenjie Zhou, Xuanmo Zhang and Yingguo Gao.

%%%%%%%%%%%%%%%%%%%%%%%%%%%%%%%%%%%%%%%%%%%%%%%%%%%%%%%%%%%%%%%%%%%%%%%%%%%%%%%%
\bibliographystyle{ieeetr}  
\bibliography{ref} 

%\begin{thebibliography}{99}
%\bibitem{c1}
%G. Saini, R. O. Panicker, B. Soman and J. Rajan, "A comparative study of different auto-focus methods for mycobacterium tuberculosis detection from brightfield microscopic images," 2016 IEEE Distributed Computing, VLSI, Electrical Circuits and Robotics (DISCOVER), Mangalore, 2016, pp. 95-100.
% J.G.F. Francis, The QR Transformation I, {\it Comput. J.}, vol. 4, 1961, pp 265-271.

% \bibitem{c2}
% H. Kwakernaak and R. Sivan, {\it Modern Signals and Systems}, Prentice Hall, Englewood Cliffs, NJ; 1991.

% \bibitem{c3}
% D. Boley and R. Maier, "A Parallel QR Algorithm for the Non-Symmetric Eigenvalue Algorithm", {\it in Third SIAM Conference on Applied Linear Algebra}, Madison, WI, 1988, pp. A20.

%\end{thebibliography}

\end{document}